\documentclass[runningheads]{llncs}
\usepackage{makecell}
\usepackage{graphicx}
\usepackage{color}
\usepackage{nicefrac}

\newif\ifdraft
\draftfalse
\drafttrue 

\definecolor{orange}{rgb}{1,0.5,0}
\definecolor{violet}{RGB}{70,0,170}
\definecolor{magenta}{RGB}{170,0,170}
\definecolor{dgreen}{RGB}{0,150,0}

\usepackage[normalem]{ulem}

\ifdraft
\newcommand{\PF}[1]{{\color{red}{\bf PF: #1}}}

\newcommand{\PUW}[1]{{\color{blue}{\bf PUW: #1}}}

\newcommand{\MS}[1]{{\color{dgreen}{\bf MS: #1}}}

\else
\newcommand{\PF}[1]{}

\newcommand{\PUW}[1]{}

\newcommand{\MS}[1]{}

\fi

\newcommand{\comment}[1]{}
\newcommand{\parag}[1]{\vspace{-3mm}\paragraph{#1}}

\newcommand{\gE}{\mathcal{E}} 

\newcommand{\mY}{\mathbf{Y}} 
\newcommand{\mU}{\mathbf{U}}   

\newcommand{\unet}[0]{\textbf{U-NET}}
\newcommand{\vnet}[0]{\textbf{V-NET}}
\newcommand{\ternausnet}[0]{\textbf{TernausNet}}
\newcommand{\linknet}[0]{\textbf{LinkNet34}}
\newcommand{\resnet}[0]{\textbf{ResNet50}}
\newcommand{\resnetse}[0]{\textbf{ResNet50-SE}}

\newcommand{\voxmesh}[0]{\textbf{Voxel2Mesh}}
\newcommand{\pixmesh}[0]{\textbf{Pixel2Mesh-3D}}
\newcommand{\pixelmesh}[0]{\textbf{Pixel2Mesh-3D}}

\newcommand{\fv}[0]{\textbf{x}}
\newcommand{\fy}[0]{\textbf{y}}
\newcommand{\ff}[0]{\textbf{f}}
\newcommand{\fm}[0]{\textbf{z}}
\newcommand{\bv}[0]{\mathbf{v}}

\usepackage{multirow}

\begin{document}
\title{Voxel2Mesh: 3D Mesh Model Generation from Volumetric Data}

\author{Udaranga Wickramasinghe\inst{1} \and
	Edoardo Remelli\inst{1} \and \\
	Graham Knott\inst{2} \and 
	Pascal Fua\inst{1}}

\authorrunning{U. Wickramasinghe et al.}
%
\institute{
	Computer Vision Laboratory, \'Ecole Polytechnique F\'ed\'erale de Lausanne, Switzerland \\
	\email{udaranga.wickramasinghe@epfl.ch}
	\and BioEM Laboratory, 
	\'Ecole Polytechnique F\'ed\'erale de Lausanne, Switzerland}

%
%
%
%
\maketitle              
%

%
%

\begin{abstract}

	CNN-based volumetric methods that label individual voxels now dominate the field of biomedical segmentation. However, 3D surface representations are often required for proper analysis. They can be obtain- ed by post-processing the labeled volumes which typically introduces artifacts and prevents end-to-end training. In this paper,  we therefore introduce a novel architecture that  goes directly from 3D image volumes to 3D surfaces without post-processing and with better accuracy than current methods. We evaluate it on Electron Microscopy and MRI brain images as well as CT liver scans. We will show that it  outperforms state-of-the-art segmentation methods.
		
\end{abstract}

\comment{ In this paper, we show that simultaneously performing the segmentation and recovering a 3D mesh that models the surface can boost performance. 
	
\begin{flushright}
	To this end, we propose an end-to-end trainable two-stream encoder/decoder architecture. It comprises a single encoder and two decoders, one that labels voxels and the other outputs the mesh. The key to success is that the two decoders communicate with each other and help each other learn. This goes beyond the well-known fact that training a deep network to perform two different tasks improves its performance. 

We will demonstrate substantial performance increases on two very different and challenging datasets.  
\end{flushright}
}

\vspace{-5mm}
\section{Introduction}

State-of-the-Art volumetric segmentation techniques rely on Convolutional Neu-ral Networks (CNNs) operating on an image volume~\cite{Cicek16,Shvets18,Milletari16}. However in clinical and research practice, a mesh representation is often required to model the surface morphology and to compute area-based statistics. Unfortunately, conver- ting volumes into surfaces  relies on algorithms such as Marching Cubes~\cite{Lorensen87} followed by mesh smoothing, which is not differentiable, prevents end-to-end training, and introduces artifacts.


\begin{figure*}
\centering
\includegraphics[height=9.5cm]{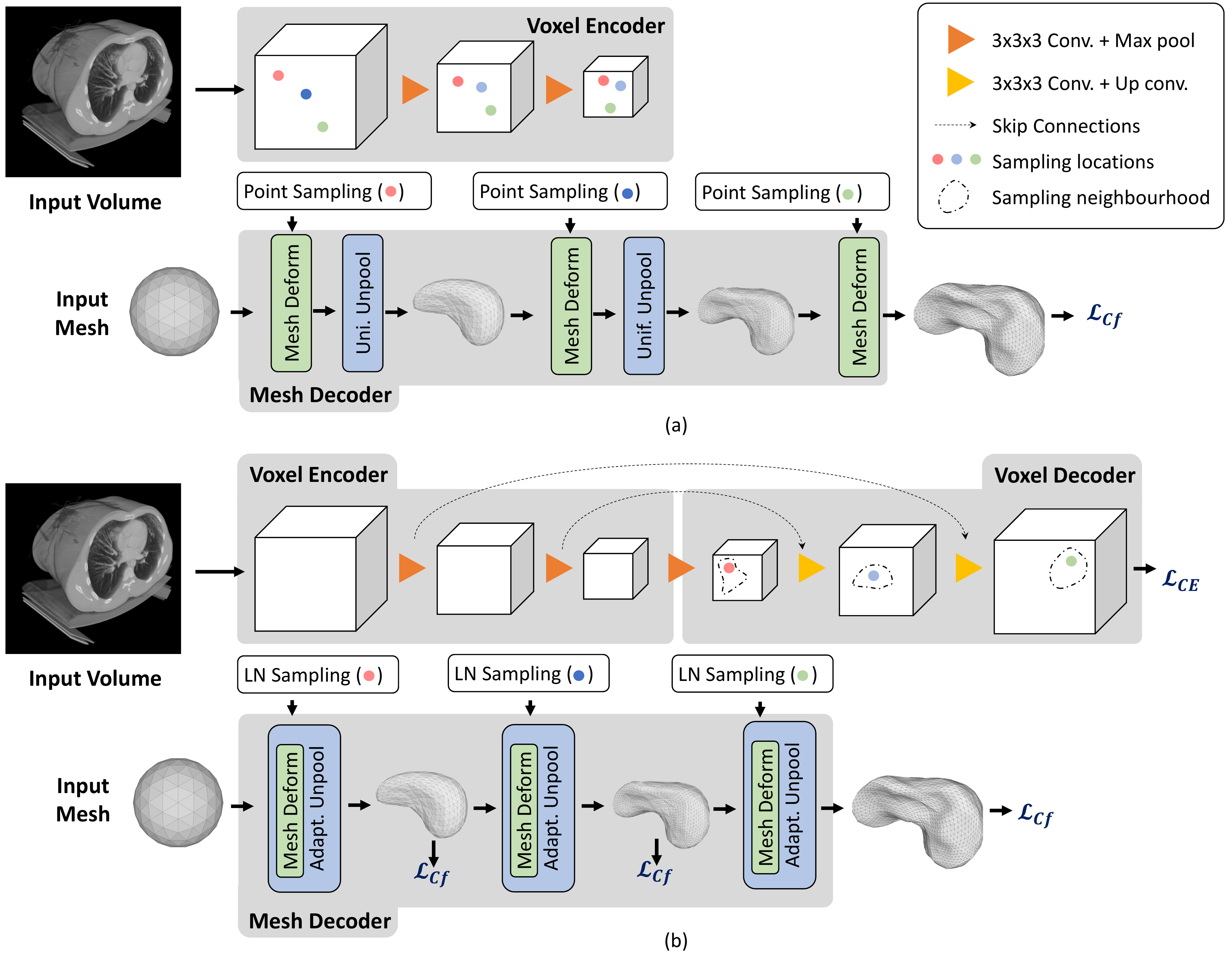}
\vspace{-8mm}
\caption{{\bf Architectures} {\small (a) The \pixmesh{} architecture, a straightforward extension of~\cite{Wang18},  uses a surface decoder but no voxel decoder. (b) By contrast, our \voxmesh{} architecture takes as input an image and spherical mesh. They are jointly encoded and then decoded into cubes and meshes of increasing resolution. At each mesh decoding stage, the decoder first receives as input the current mesh and a set of features  sampled from the cube of corresponding resolution. Then the mesh is deformed and refined non-uniformly by adding vertices only where they are needed.  }}
\vspace{-4mm}
\label{fig:arch}
\end{figure*}

We therefore introduce an end-to-end trainable architecture that goes directly from volumetric images to 3D surface meshes. Our \voxmesh{} architecture is depicted by Fig.~\ref{fig:arch} (b). It comprises a voxel encoder, voxel decoder and a mesh decoder. The two decoders communicate at all resolution levels and our approach incorporates two innovative features that are key to performance.
\begin{itemize}

 	\item {\bf Learned Neighborhood Sampling.} The mesh decoder learns to sample the output of the volume decoder only where needed, that is, in the  neighbor- hood of output vertices.
	
	\item {\bf Adaptive Mesh Unpooling.} Accurately representing the surface requires densely sampled mesh vertices in high-curvature areas but not elsewhere. We introduce an adaptative mesh unpooling scheme that achieves this results and eliminates the need for exponentially large amounts of memory that uniform unpooling requires. 
	
\end{itemize}
Our contribution therefore is a novel architecture that takes a 3D volume as input and yields an accurate 3D surface without any post-processing. We evaluate it on Electron Microscopy and MRI brain images as well as CT liver scans. We will show that it  outperforms state-of-the-art segmentation methods, especially when the training set is relatively small.  


\begin{figure*}
\centering
\includegraphics[height=4.05cm]{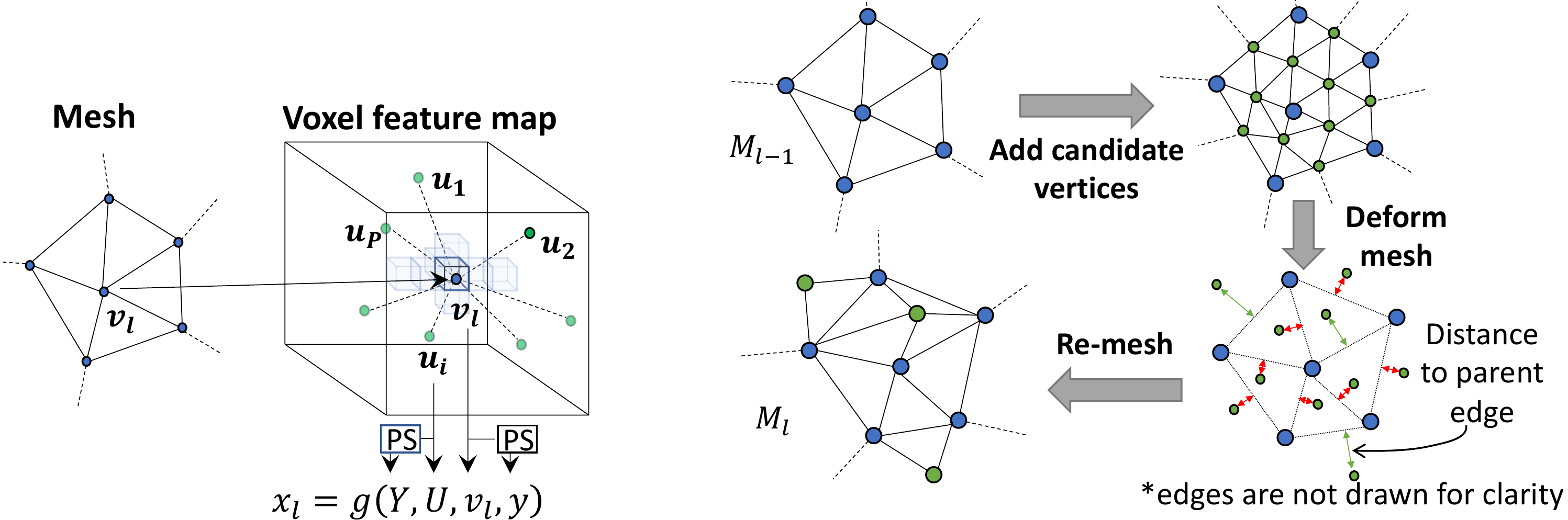}
\vspace{-7mm}
\caption{\small {\bf Approach.} \hspace{-1.5mm} (a) \hspace{-1mm} Learned Neighborhood Sampling. (b) \hspace{-1mm} Adaptive Mesh Unpooling.}
\vspace{-5mm}
\label{fig:sub_modules}
\end{figure*}






\vspace{-2mm}
\section{Related Work} 
\vspace{-1mm}
\label{sec:related}

CNN-based volumetric methods such as U-Net and its variants ~\cite{Cicek16,Milletari16,Shvets18,Iglovikov18} now dominate  biomedical image segmentation. This is evident from the CHAOS challenge~\cite{Kavur19} and Medical Segmentation Decathlon~\cite{IsenseeEmail18} results. The winners  of both competitions used ensembles of methods relying on volumetric CNNs. 

\vspace{-4mm}
\subsection{Importance of Surface Models}
 
Many biological imaging tools are designed to study the morphology of structures such as cells, organs, or tissues. Researchers usually prefer to visualize them as 3D surfaces at the required level of detail everywhere and are not limited by the voxels resolution. Even though volumes and distances can be estimated using voxels, analyzing surfaces is best accomplished using meshes.


But, state-of-the-art methods produce volumetric descriptions and therefore must be converted to surface meshes. This conversion typically relies on algo- rithms such as Marching Cubes~\cite{Lorensen87} followed by mesh smoothing. This introduces artifacts and prevents end-to-end training. We now turn to existing approaches that mitigate these difficulties. 

 \vspace{-4mm}
 \subsection{Deformable Models}
 
Deformable surface models became popular in the 1990s to model biological structures in volumetric data~\cite{McInerney96,He08a} and are still being developped~\cite{Leventon00,Prevost13,Jorstad14a}. They are now used in conjunction with deep networks~\cite{Marcos18,Dong18} trained to return the energy function the deformed models should minimize. 

While they can remove some of the artifacts introduced by converting volumes to surfaces, their use makes the processing pipeline more complex and still prevents end-to-end training. Furthermore, they are not  well-suited to segmenting structures that exhibit high inter-sample shape variations, such as the synaptic junctions used in our experiments.

\vspace{-4mm}
\subsection{Deep Surfaces}

In this context, the Pixel2Mesh~\cite{Wang18} approach and its more recent variants~\cite{Pan19,Wen19}  are of particular interest. They are among the very few approaches that go {\it directly} from 2D images to 3D surface meshes without resorting to an intermediate stage. They take an image and encodes it into a set of progressively smaller feature maps. This set is then used at each stage of the decoding process to produce an increasingly accurate mesh. This approach is easily extended to handle 3D image volumes using the architecture depicted by Fig.~\ref{fig:arch}(a) and we will refer to this extension as \pixmesh{}. Unfortunately, as we will see, it was conceived for a different purpose and its design choices makes it suboptimal for handling 3D image volumes.



 
\section{Method}
Our \voxmesh{} architecture is depicted by Fig.~\ref{fig:arch}(b). It takes an image volume as input and returns a 3D surface mesh. The image volume is first encoded into smaller latent volumes that serves as input to the {\it voxel decoder}. Then the voxel decoder generates a pyramid of cubes of increasing resolution whose voxels contain feature vectors. Finally the \textit{mesh decoder} generates increasingly precise deformations of an initial spherical mesh using feature vectors extracted from voxel decoder. The voxel encoder and voxel decoder pictured at the top of Fig.~\ref{fig:arch}(b) are based on a standard U-Net \cite{Cicek16} architecture.  

A key specificity of our approach is that both the sampling of the voxel features and the location and number of the new vertices needed to refine the mesh are adaptive so that the final mesh is refined where it needs to be, and only there.

\vspace{-4mm}
\subsection{Mesh Decoder} 
\label{sec:meshDecode}

The input to the mesh decoder is a sphere mesh with 3D vertices forming facets whose edges we use to perform graph convolutions.  It learns to iteratively refine the sphere mesh to match the target object.  

Let us denote by $l=0$ the input to the decoders and by $1 \leq l \leq L $ the output of subsequent blocks. For each mesh vertex, we write 
\begin{equation}
\label{eq:sampling}
\fm_{l}   =  h_l(\fv_{l}, \fm_{l-1},\bv_{l-1}) \mbox { and } 
\bv_{l}   =  \bv_{l-1} + \Delta_{l}(\fm_{l}) \; ,
\end{equation}
%
%
where $\textbf{v}_{l}$ are the 3D vertex coordinates after block $l$; $\fv_l$ and $\fm_{l-1}$ are the feature vectors produced by blocks $l$ and $l-1$ in the voxel and mesh decoder, respectively; $h_l$ and $\Delta_l$ are two functions implemented by 4 graph convolution layers each, whose weight we learn during training. By convention, we take $\fm_{0}$ to be an empty feature vector.  We write the graph convolutions as 

\begin{equation}
\label{eq:graph_conv_equation}
{\ff{~'}} =  {w}_1\ff+ \frac{1}{|\mathcal{N}(\bv_l)|} \sum_{\bv^{i}_l \in \mathcal{N}(\bv_l)  }\ff^{~i}{w}_2 e^{-\nicefrac{d_i^2}{\sigma^2}}\; ,
\end{equation}
\noindent 
where $\ff'$ and $\ff$ are the feature vector associated with the vertex $\bv_l$ before and after the convolution. $\mathcal{N}(\bv_l)$ is the set of neighborhood vertices of $\bv_l$ and $\fy^{~i}$  is the feature vector corresponding to the neighboring vertex $\bv^i_l$. $d_i = \sqrt{|| \bv^i_l - \bv_l ||}$. $\textbf{w}_1, \textbf{w}_2$ and $ \sigma$ are  weights learned during training. 

\vspace{-3mm}
\subsubsection{Learned Neighborhood Sampling (LNS)}
\label{sec:sampling}

The feature vector $x_l$ in Eq.~\ref{eq:sampling} is extracted from voxel features by feature sampling at locations that are functions of the mesh vertices. Since voxel features lie on a discrete grid, we use tri-linear interpolation to sample features and refer to this as \textit{point sampling}.  Current approaches only sample at exact vertex locations \cite{Wang18} or in pre-determined neighborhoods around the vertex \cite{Wen19}. This restrict the sampler's ability to pool information from its neighborhood. 

Instead, we introduce our LNS strategy that learns optimum sampling loca- tions. It first samples feature vector $\fy$ at a given vertex  $\bv_l$ using point sampling. We then train a neural function to return the set of neighborhood points to sample
\begin{equation} 
 \mU =  \{u_i\}^P_{i=1} = f (\fy, \bv_l) \; .
\end{equation}
Next, the set of features $ \mY=\{\fy_i\}^P_{i=1}$ are sampled at $ \{u_i \}^P_{i=1}$, again using point sampling. Finally, the feature vector $\fv_l$ corresponding to vertex $\bv_l$ is given by another neural function 
\begin{equation} 
\fv_l = g(\mY, \mU, \fy, \bv_l) \; .
\end{equation}
As shown in Fig.~\ref{fig:sub_modules} (a),  LNS only samples from the voxel feature map corres- ponding to the mesh deformation stage. This is in contrast to earlier sampling strategies, that samples from all feature maps at each stage which yields graph convolution networks with many more weights in their mesh deforming module and thus over-fitting when trained with smaller datasets.

\vspace{-4mm}
\subsubsection{Adaptive Mesh Unpooling} 
\label{sec:unpool}

High accuracy requires enough vertices to pro- perly fit the underlying surface. We could  start with a sphere with many vertices but this is neither computationally nor memory efficient. Therefore, \pixmesh{} and its variants use an uniform unpooling strategy that gradually increase the vertex count. Unfortunately, the vertex count still increases exponentially.

To prevent this, we introduce the adaptive unpooling strategy depicted by Fig.~\ref{fig:sub_modules} (b). First, we add candidate vertices using uniform unpooling strategy. Then the mesh is deformed and we compute the shortest distance from each candidate vertex to its parent edge as indicated by red and green arrows in Fig.~\ref{fig:sub_modules} (b). If the distance is greater than a threshold, we keep them, otherwise we discard them. This looses edge connectivity making re-meshing necessary. To this end, we exploit the fact that the mesh decoder learns a continuous mapping from the surface points on the input sphere to those on the object surface. This relationship enables us to find the corresponding points on the sphere's surface for each mesh vertex. We compute the convex hull to restore edge connectivity $\gE'$ between the points on the sphere. $\gE'$ can then be directly transferred to the target mesh because the mapping is continuous.  Since our remeshing only recomputes the neighbors of each vertex and does not perform any non-differentiable operations on vertex variables, it preserves overall differen- tiability.


\subsection{Loss function}

We use the cross entropy loss $\mathcal{L}_{ce}$ with ground-truth volumes and the Chamfer distance $\mathcal{L}_{cf}$  to points at the boundary of the same ground-truth volumes to train the voxel and mesh decoders, respectively. Instead of using mesh vertices when evaluating $\mathcal{L}_{cf}$, we randomly sample points from the 3D mesh \cite{Funkhouser02}. We also introduce three regularization terms  normal loss $\mathcal{L}_{n}$, laplacian loss $\mathcal{L}_{lap}$, and edge length loss $\mathcal{L}_{el}$  to improve convergence and smooth the output mesh \cite{Wang18}. We write the complete loss as 
\begin{equation}
\mathcal{L} = \sum_{l=1}^{L}\mathcal{L}_{cf}^l  +  \lambda_1\mathcal{L}_{ce} + \lambda_2\mathcal{L}_{n} + \lambda_3\mathcal{L}_{lap} + \lambda_4\mathcal{L}_{el} \; ,
\end{equation}
where $L$ is the number of stages in the mesh decoder.

\section{Experiments}

\vspace{-1mm}
\subsection{Datasets}

\noindent We  tested our approach on 3 datasets. We describe them below briefly and provide more details on the training and testing splits in the supplementary material.

\begin{figure*}
\centering
\includegraphics[height=7.5cm]{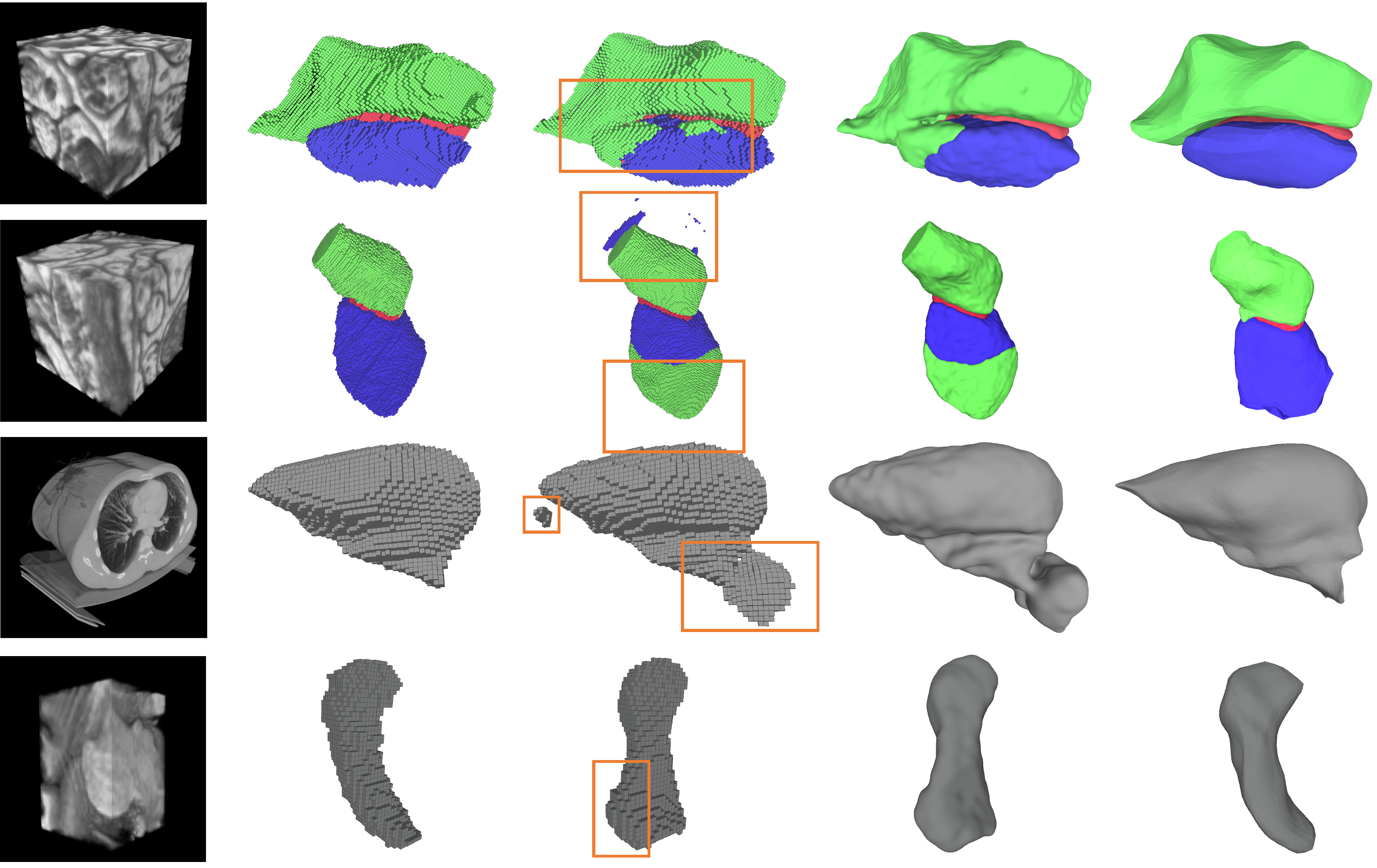}\\
(a) \hspace{2cm}(b) \hspace{2cm}(c) \hspace{2cm}(d)\hspace{2cm}(e)
\vspace{-4mm}
\caption{{\bf Qualitative results.} {\small (a) Input volumes. EM (row 1,2), CT(row 3), MRI(row 4) (b) Ground truth (c) CNN baseline (d) CNN baseline + post processing (e) \voxmesh{}. The orange boxes highlight false positive regions. }}
\label{fig:results}
\end{figure*}

\parag{Synaptic Junction Dataset.} It comprises a $500 \times 500 \times 200$ FIB-SEM image stack of a mouse cortex. We extracted 13 volumes roughly centered around a synapse for training and 13 for testing. The task is to segment the pre-synaptic region, post-synaptic region, and synaptic cleft. They are shown in blue, green, and red respectively in the first two rows of Fig.~\ref{fig:results}.

\parag{Hippocampus Dataset.} It consists of  260 labeled MRI image cubes from the Medical Segmentation Decathlon \cite{hippo19}.  The task is to segment the liver, as depicted by the third row of Fig. \ref{fig:results}. 

\parag{Liver dataset.} It consists of 20 labeled CT image cubes from the CHAOS challenge \cite{Kavur19}. The task is to segment the hippocampus as shown in the last row of Fig. \ref{fig:results}. 

\begin{figure*}
\centering
\includegraphics[height=3.05cm]{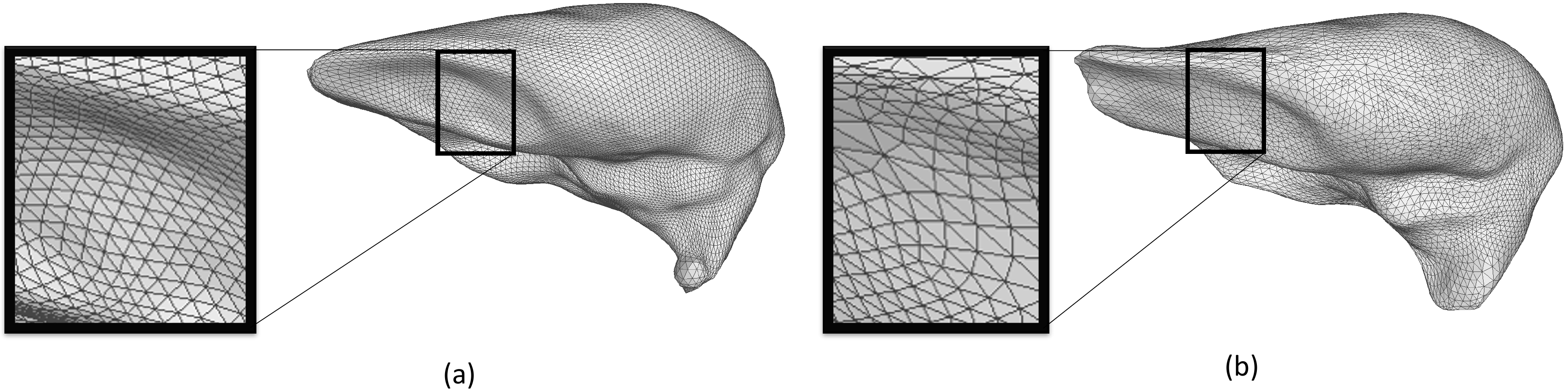}
\vspace{-8mm}
\caption{{\bf Levels of resolution.} {\small (a) \pixmesh{} result (10422 vertices). (b) \voxmesh{} result (7498 vertices). With our adaptative unpooling, we obtain better results with fewer vertices. }}
\label{fig:vcount}
\end{figure*}
 

\begin{table*}[htbp]
\begin{small}
\centering
\caption{\small Comparative results on three datasets using the IoU metric.}
\label{table:main_results}
\vspace{-3mm}
\begin{tabular}{|l|c|c|c|c|c|}
	\hline
	\multicolumn{1}{|c|}{\multirow{2}{*}{~}} & \multirow{2}{*}{Liver}        & \multirow{2}{*}{Hippo.} & \multicolumn{3}{c|}{Synaptic Junction}            \\ \cline{4-6} 
	\multicolumn{1}{|c|}{}                        &                               &                              & Pre-Synap.             & Synapse       & Post-Synap.           \\ \hline
	\ternausnet{}     \cite{Iglovikov18}           & \textit{84.4 $\pm$ 1.3 }              & 78.4 $\pm$ 1.2               & 73.5 $ \pm$ 1.3 & 64.4 $\pm$ 0.5 				& \textit{78.4 $\pm$ 1.3} \\ \hline
	\linknet{}     \cite{Shvets18}                & 82.8 $\pm$ 1.4                & 79.4 $\pm$ 0.8               & 72.3 $ \pm$ 0.5 			& 63.2 $\pm$ 1.2 			& {78.2 $\pm$ 1.1} \\ \hline
	\resnet{}     \cite{Kavur19b}                 & 82.1 $\pm$ 0.7                & 80.7 $\pm$ 0.2               & 70.3 $\pm$  0.8 			& 63.3 $\pm$ 0.6 			& 76.2 $\pm$ 1.4 \\ \hline
	\resnetse{}  \cite{Kavur19b}                  & 82.6 $\pm$ 1.2                & 80.5 $\pm$ 1.3               & 71.3 $\pm$ 0.6  			& 63.6 $\pm$ 0.7 			& 76.3 $\pm$ 0.9 \\ \hline
	\vnet{}      \cite{Milletari16}                  & 81.5 $\pm$ 1.4                & 75.3 $\pm$ 1.4               & 64.3 $\pm$ 0.7  			& 65.2 $\pm$ 1.3 			& 74.1 $\pm$ 0.7 \\ \hline
	\unet{}     \cite{Cicek16}                   & 84.2 $\pm$ 1.6                & \textit{80.9 $\pm$ 1.5 }     & \textit{73.6 $\pm$ 1.3 } 	& \textit{67.2 $\pm$ 0.8 }	& {78.2 $\pm$ 0.9} \\ \Xhline{3\arrayrulewidth}
	\textit{Best CNN} + \textbf{CLN}    & 84.6 $\pm$ 1.7          & 81.1 $\pm$ 1.5              & 74.5 $\pm$ 1.2  			& {67.6 $\pm$ 0.8} 	& 79.5 $\pm$ 0.9 \\ \hline
	\textit{Best CNN} + \textbf{FPP}       & 84.3 $\pm$ 1.7          & 80.8 $\pm$ 1.5              & 74.2 $\pm$ 1.2  			& \textbf{67.4 $\pm$ 0.8} 	& 79.3 $\pm$ 0.9 \\ \Xhline{3\arrayrulewidth}
	\voxmesh{}                   & \textbf{86.9 $\pm$ 1.1} & \textbf{82.3 $\pm$ 0.9}               	  & \textbf{77.3 $\pm$ 1.2}  	& 65.3 $\pm$ 1.2 			& \textbf{83.2 $\pm$ 1.6} \\ \hline
\end{tabular}
\end{small}
\vspace{-3mm}
\end{table*}


\vspace{-4mm}
\subsection{Baselines}

As the architecture of \voxmesh{} borrows from  \unet{} and \pixelmesh{}, they both constitute natural baselines. As state-of-the-art CNN based appro- aches, we use \ternausnet{}, \linknet{}, \resnet{} and \resnetse{}, which belong to the ensemble of architectures that won the CHAOS challenge. \unet{} was used as the base architecture by the winner of Medical Segmentation Decathlon. We also use \vnet{},  a widely used variant of \unet{}, as a baseline. Since we are working with volumetric data, we use 3D variants of all these architectures. 


\vspace{-5mm}
\subsection{Comparative Results}

Figs.~\ref{fig:results} and Figs.~\ref{fig:vcount}  depict our results qualitatively and we report quantitative ones in Tab. \ref{table:main_results}. As all the baselines shown above the first thick line return volumetric descriptions, we rasterize the mesh that \voxmesh{} returns and use intersection over union (IoU) score to compare all the results against the ground truth. In all cases, we trained the networks three times with different initializations and we report the mean IoU and its standard deviation.

For completeness, we simulated a post-processing pipeline by selecting the best performing baseline for each of the three datasets, removing small false positive regions outside the object by connected component analysis, running Marching Cubes followed by smoothing using Algebraic Point Set Surfaces~\cite{Guennebaud07}. We refer to the result obtained by only removing the false positives as \textit{Best CNN} + \textbf{CLN}  and the one with full post-processing as \textit{Best CNN} + \textbf{FPP}.

\voxmesh{} performs best in all cases except the synaptic cleft, where it comes second. This can ascribed to the fact that synaptic clefts sometimes have holes in them, which a spherical mesh cannot capture. The improvement is most significant in the two datasets---liver and synaptic junction---with fewer training samples compared to the hippocampus dataset that features more training data.

\vspace{-1mm}
\subsection{Ablation Study}

\pixelmesh{} is the baseline closest to our approach because it directly outputs a 3D surface. Point Sampling (\textbf{PS}) and Uniform Mesh Unpooling (\textbf{UMU}) are the two modules in that play the same role as our Learned Neighborhood Sampling (\textbf{LNS}) and  adoptive mesh unpooling (\textbf{AMU}).  To check the importance of these strategies, we replaced them in our \voxmesh{} pipeline by the equi- valent ones in \pixmesh{}. We also evaluate the performance of Hypothesis Sampling (\textbf{\textbf{HS}}), the sampling strategy of~\cite{Wen19} that relies on a fixed neighborhood sampling. We report the results in 
Tab~\ref{table:mesh_results} in Chamfer distance terms. They  demonstrate that our adaptative strategies \textbf{LNS}+\textbf{AMU} deliver a clear benefit. 


\begin{table}[ht!]
       \centering
       \caption{\small Comparative results against CNN based mesh deforming baselines.  } 
       \label{table:mesh_results}
       \vspace{-3mm}
       \begin{small}
	\begin{tabular}{|l|c|c|c|c|}
		\hline
		\multirow{2}{*}{~}          & \multicolumn{2}{c|}{Liver}            & \multicolumn{2}{c|}{Hippocampus}      \\ \cline{2-5} 
		& IoU            & Cf.                  & IoU            & Cf.                  \\ \hline
		\textbf{PS}  + \textbf{UMU} 		  				                 & 83.3 $\pm$ 0.8 & 3.3 $\times 10^{-3}$ & 78.8 $\pm$ 1.1  & 2.9 $\times 10^{-3}$ \\ \hline
		\textbf{HS}   + \textbf{UMU}   				  & 84.2 $\pm$ 0.6 & 2.8 $\times 10^{-3}$ & 79.9 $\pm$ 0.9 & 2.3 $\times 10^{-3}$ \\ \hline
		\textbf{LNS} 	+ \textbf{UMU} 							     & 85.6 $\pm$ 0.9 & 2.1 $\times 10^{-3}$ & 81.2 $\pm$ 1.2  & 1.8 $\times 10^{-3}$ \\ \Xhline{3\arrayrulewidth}
		\textbf{LNS} + \textbf{AMU}  (\voxmesh{})& \textbf{86.9 $\pm$  1.1}   & \textbf{1.3 \boldmath$\times 10^{-3}$} & \textbf{82.3 $\pm$ 0.9} & \textbf{1.1 \boldmath$\times 10^{-3}$} \\ \Xhline{3\arrayrulewidth}
	\end{tabular}
	\vspace{-5mm}
\end{small}

\end{table}



\section{Conclusion}

We have proposed an end-to-end trainable architecture that takes an image volume as input and outputs a 3D surface mesh. This makes the post processing steps usually required to obtain such a mesh from a volumetric representation unnecessary, while preserving accuracy. Our adaptative sampling and unpooling strategies are key to this result. Not only does our architecture deliver good results, it also bridges the gap between voxel-based and surface-based represen- tations. In future work, we plan to extend our approach to structures with more complex topologies, such as the synaptic cleft and its potential holes.


\comment{
We have proposed an end-to-end trainable two stream encoder-decoder architecture that simultaneously produces volumetric and surface descriptions. Crucially, this is more than two decoders working in parallel on a joint latent representation as in many multi-task architectures: The voxel decoder provides features to the mesh decoder at every step, which guarantees that these features are useful for both segmentation and surface reconstruction. 

We have demonstrated that this joint architecture performs better than either decoder alone and than the baselines. This confirms the importance of continued surface modeling in biomedical image segmentation. A limitation of the current approach is that it can only handle surfaces of genus 0. In future work, we will explore more sophisticated architectures that can remove this limitation. 
}
%
%

\bibliographystyle{splncs04}
\bibliography{string,vision,biomed} 

\end{document}